# The role of time in considering collections

Françoise Gayral, Daniel Kayser and François Lévy
LIPN Institut Galilée (UMR 7030 du C.N.R.S.)
99 Avenue Jean-Baptiste Clément
F-93430 Villetaneuse
{fg,dk,fl}@lipn.univ-paris13.fr

**The role of time in considering collections**

We are working on the understanding of plurals in the framework of Artificial Intelligence. In the large literature concerned by this topic, the emphasis is essentially put on the contrast between collective, distributive and cumulative readings with classical examples such as *Paul and Mary lifted the piano, The students are young, 3 girls ate 4 sandwiches*. A number of sophisticated logical theories have been constructed to deal with these readings and to embed both individuals corresponding to singular terms, and collections corresponding to plural terms. Two directions have been adopted. The first one (Hausser 74) (Bennett 75) considers collections as higher-order individuals; the other makes no ontological distinction between the two types of objects but discriminates them in the interpretation domain which is enriched with non-atomic individuals and with a partial order induced by a 'part-of' relation between a non-atomic individual and the individuals which compose it (Scha 81) (Link 84) (Kamp & al 93) (Landman 89).

We carry on a corpus analysis (texts extracted from the newspaper *Le Monde*). This study shows that these theories, in paying too much attention to the distributive/collective contrast, focus on questions which are not necessarily relevant, in particular concerning the cumulative reading (asking for the distribution of the protesters on the entrances in *many protesters blocked the entrances of the town* is inappropriate) and miss many other phenomena concerning plural. The most neglected issue is the role of time. Since interpreting plurals needs obviously the construction of collection(s), the evolution of the collection across time is often crucial and has to be accounted for. This phenomenon is particular salient in sentences whose interpretation requires the plural nominal phrases of the sentence to be evaluated in several situations (Enc 86). In (Gayral & al 2001), we have contrasted a *de dicto* collection where the collection can be considered as persisting over these situations even if its members change as in (2) with a *de re* collection whose composition does not vary through time (1), (3) remaining ambiguous -the choice between *de re* and *de dicto* depending on the context of enunciation-.

1. *Pendant plusieurs heures, les aborigènes ont fait un sitting pour protester contre leur sort (For several hours, the aborigines remained seated to protest against their lot)*
2. *Les étudiants des classes préparatoires sont de moins en moins issus de milieux défavorisés (Selective classes students are coming less and less from lower classes)*
3. *Les jeunes consomment un peu moins de tabac et plus de cannabis (Young people consume a little less tobacco and more cannabis)*

The interference between time and plural is the main object of this communication. Focusing on plural which have to be evaluated at more than one time point, we show that many sources of knowledge collaborate in order to get the *de re* or *de dicto* interpretation and we propose a formalism which deals with this distinction. Here is a brief survey.

**Elements for an analysis**

We consider that by default, the interpretation is *de re*. The *de dicto* interpretation is forced when some constraints are at play. Here are some of them. We will develop others in the long paper.

- The plural nominal phrase is argument of a comparative or evolutive predication in which the comparison or the evolution cannot concern the same elements.

This impossibility can be given by intrinsic properties of the predication, as in (2). The predication 'coming from lower classes', although typically distributive, denote a property which cannot change over time for a given individual. No other "collective interpretation" of the predicate being available, the collection is *de dicto* and is interpreted in terms of ratio. There is a decrease of the relative part of the lower classes students among the population considered.

This impossibility can result from the context as when (3) is found in a survey carried out each year during the military service day that 18-years old people must attend. The young people under consideration cannot be the same from year to year, and the interpretation is forced to a *de dicto* collection with (at least) two instantiations, each of them being partitioned in two subsets (the cannabis smokers and the tobacco smokers); the comparison of the four obtained ratios gives rise to the affirmation.

In another context, where the young people form a well-defined set (replace, for example, *young people* by *Paul's friends*), the collection would be *de re* and two interpretations are available; the first one corresponds to an individual evolution of each of them since the smoking activity can evolve for a given person through time, the other one corresponds to the estimation of the global consumption (of tobacco and of cannabis) for the same individuals taken at two different moments.

- There is a mismatch between temporal properties which are either expressed in different parts of the sentence or are part of more general knowledge. For lack of place, we just contrast (4) contrast with (1) where the human life span forbids to speak about the same people.

4. *For several centuries, the aborigines remained ignored*

**Elements for a formalism**

We choose a notation which reminds the stages of individuals advocated by (Carlson 80) and which allows to denote "temporal slices" of entities, atomic or not. Here are some details:

– $\alpha@t$ denotes the "temporal slice" of an entity $\alpha$ at/during the temporal reference $t$[1], $\alpha$ having a life span $ls(\alpha)$ of which $t$ is a part[2].

– $S \equiv \mathsf{P}(x,\_,z)$ denotes in extension the collection of entities which satisfy in the appropriate position the predicate $\mathsf{P}$, i.e. $\{y \mid \mathsf{P}(x,y,z)\}$.

– a *de dicto* set, which exists independently of time as for Carlson's individual, is noted:
$S_{dicto} \equiv \mathsf{P}(\_)$[3], and $S$ can have different realizations at different moments, i.e:
for all $t$: $S@t = \{ x@t \mid \mathsf{P}(x@t) \}$ which represent "temporal slices" for the set S.

For example, if we consider the first part of (3) (*Young people consume less tobacco*) in its first interpretation, we need :

- the *de dicto* collection $Y_{dicto} \equiv$ 18-old (\_) with two instantiations
$Y@2002 = \{ x@2002 \mid$ 18-old ($x@2002$) $\}$ and $Y@2003 = \{ x@2003 \mid$ 18-old ($x@2003$) $\}$, if the two situations take place in 2002 and 2003.

- for each instantiation, the sub-collection of those who smoke tobacco : $Yt@2002 = \{ x@2002 \mid x@2002 \in Y@2002 \land \mathrm{Smoke}(x@2002, tobacco)\}$, idem for $Yt@2003$

This reading says that card($Yt@2003$)/card($Y@2003$) < card($Yt@2002$)/card($Y@2002$).

– On the contrary, a *de re* set is necessarily attached to time, the moment under consideration being noted in the subscript of the following notation:
$S_{re} \equiv_t \mathsf{P}(\_)$. S is here the collection of all individuals $x$ that have property $\mathsf{P}$ at the common time reference $t$. This collection can be considered at any other moment; so, for all $t'$, we can define:
$S@t' = \{ x@t' \mid \mathsf{P}(x@t) \}$. This allow us to represent a collection grouping individuals that shared at a given time a given property, even if the sentence considers them at a moment where they do not possess it any more or not yet.

The two other interpretations of (3) require:

- the *de re* collection of Paul's friends: $F_{re} \equiv_t \{x \mid \mathrm{Friend}(x, Paul)\}$
- the *de dicto* collection of their tobacco consumption with two instantiations. If $\mathrm{Cons}_y(x)$ is the quantity of substance y consumed by the individual x, we define $C_{dicto} \equiv \{\mathrm{Cons}_{tobacco}(f) \mid f \in F\}$

The individual evolution is expressed by: $\mathrm{Cons}_{tobacco}(f@2002) > \mathrm{Cons}_{tobacco}(f@2003)$ for each $f \in F$

The global evolution is expressed by: $\sum_{c \in C@2002} c > \sum_{c \in C@2003} c$

---

[1] Temporal references are implicitly handled as time points, but nothing prevents, when needed, to consider them as intervals.
[2] Some entities are perceived as invariant through time, and for them, for all $t, t'$: $\alpha@t = \alpha@t'$.
[3] For sake of simplicity and for covering all the different cases, we omit here the other arguments.